\documentclass[11pt]{article}

\usepackage[final]{acl}

\usepackage{times}
\usepackage{latexsym}

\usepackage[T1]{fontenc}

\usepackage[utf8]{inputenc}

\usepackage{microtype}

\usepackage{inconsolata}

\usepackage{graphicx}

\usepackage{xcolor}
\usepackage{listings}
\usepackage{booktabs}

\title{
Is a Document Educational or Just Wikipedia-Style? -- Pitfalls of Classifier-Based Quality Filtering
}

\author{Mateusz Klimaszewski\textsuperscript{1}, Piotr Andruszkiewicz\textsuperscript{1,2} \\
  \textsuperscript{1}Warsaw University of Technology,
  \textsuperscript{2}IDEAS Research Institute\\
        \texttt{firstname.lastname@pw.edu.pl}
        }

\begin{document}
\maketitle
\begin{abstract}
Classifier-based Quality Filtering has recently emerged as a fundamental technique in constructing pre-training corpora. The ability to deploy a single model that can replace or supplement a set of heuristics has proven effective across numerous Large Language Models. In this work, we expose a critical vulnerability in this approach by demonstrating how a straightforward Wikipedia-style reformatting operation can substantially alter a model's quality assessment and enable low-quality content to surpass filtering thresholds. Our analysis reveals that the FineWeb-Edu CQF model would reverse its filtering decision for approximately 7\% of evaluated documents, thereby admitting content into the pre-training corpus that would otherwise have been excluded.\footnote{Repository:\newline \url{https://github.com/mklimasz/cqf-pitfalls}}
\end{abstract}

\section{Introduction}

Large Language Models (LLMs) are pre-trained on massive data corpora, and the quality of these corpora is one of the main factors in achieving state-of-the-art performance. The standard elements on the majority of pipelines include a series of heuristic filters (e.g., language identification, near-deduplication, characters-to-other-signs ratio \cite{wenzek-etal-2020-ccnet,burchell-etal-2025-expanded}) that aim to improve downstream performance. An alternative to or supplement of heuristic filters is Classifier-based Quality Filtering (CQF) \cite{fineweb}. This solution offloads the task of filtering to pre-trained models that provide a quality score, which is used not only to filter the pre-training corpora but also to design a curriculum based on the scores. Nowadays, CQF is a cornerstone of many prominent datasets (Fineweb-Edu \cite{fineweb}, DCLM \cite{li2024datacomplm}, Nemotron-CC \cite{su-etal-2025-nemotron}) and LLMs (e.g. LLama 3 \cite{llama3}, EuroLLM \cite{eurollm9b}, SmolLM3 \cite{bakouch2025smollm3}), showing its usefulness not only in monolingual (mostly English-based filtering) but also multilingual scenarios \cite{eurollm9b,waldendorf-etal-2025-multilingual}.

CQF operates on the assumption that educational data enhances the performance of the LLM \cite{phi1,phi3} and aims to filter such data from raw corpora. Filtering operates through a proxy; i.e., an "educational" quality classifier is developed from synthetic annotations generated by an LLM, e.g., LLama-3-70B \cite{fineweb}. The proxy, typically a BERT-sized model, is necessary to operate on a pre-training scale and make filtering feasible due to computational overhead. The "educational" score (typically a value between 0 and 5) is assigned based on specific criteria by the LLM (see Appendix \ref{sec:edu_appendix}), and the proxy model is trained to replicate its behaviour.

\begin{figure}[t]
    \centering
    \includegraphics[width=\linewidth]{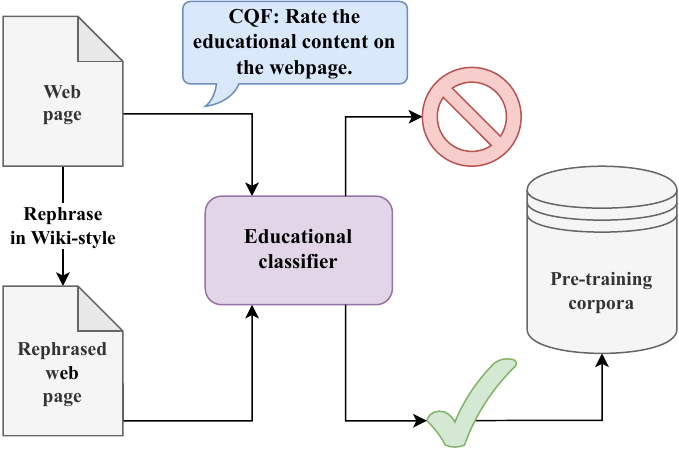}
    \caption{An educational classifier can be manipulated to assign a higher score to Wiki-style rephrased text, thereby passing the sample to the pre-training data.}
    \label{fig:teaser}
\end{figure}

In this work, we challenge the "educational" aspect of the obtained datasets and demonstrate (Sections \ref{sec:wiki} and \ref{sec:domain}) that CQFs can be Wikipedia-like or style-oriented and biased tools for filtering data, where the "educational" aspect may not be the sole outcome, as presented in Figure \ref{fig:teaser}. Our findings align with the related work of \citet{apple_cqf}, which highlights the limitations of CQF in between high-quality and low-quality examples.


The potential effects of CQF pitfalls should not be overlooked. Unlike instruction data, which is heavily filtered and, at times, manually verified, the pre-training scale makes the data, to some degree, unverifiable. Malicious actors could leverage these limitations to bypass filtering and promote their agenda by manipulating web corpora \cite{10646610}. Moreover, recent studies have shown that a precisely injected backdoor can lead to permanent damage that persists even after the LLM's alignment phase \cite{zhang2025persistent}.

\section{Related Work}

Model-based filtering has played a crucial role in constructing large-scale pre-training corpora \cite{wenzek-etal-2020-ccnet}. Traditional n-gram-based approaches, including FastText \cite{barrios-etal-2009-enriching} for language identification \cite{nllb2022,burchell-etal-2023-open} and KenLM \cite{heafield-2011-kenlm} for perplexity-based filtering, have played foundational roles in ensuring corpus quality and relevance. As LLMs have scaled in size and capability, BERT-based models \cite{devlin-etal-2019-bert,conneau-etal-2020-unsupervised} have emerged as comparatively ``lightweight'' alternatives, gradually replacing simpler statistical and shallow neural network methods.

This transition is especially noticeable in the Machine Translation field, where Quality Estimation (QE) models such as COMET \cite{rei-etal-2022-cometkiwi,rei-etal-2023-scaling} and MetricX \cite{juraska-etal-2024-metricx} have become established tools for filtering parallel corpora. However, despite the advances these QE models have enabled in translation quality \cite{specia-etal-2021-findings,zerva-etal-2022-findings,blain-etal-2023-findings,zerva-etal-2024-findings}, subsequent research has revealed their limitations and identified unintended biases that filtering can introduce into downstream models \cite{zaranis-etal-2025-watching}.

CQF has expanded the scope of model-based filtering beyond parallel corpora to orders of magnitude larger pre-training datasets. While CQF methods have demonstrated effectiveness in producing English-centric LLMs and have shown recent promise in multilingual applications \cite{eurollm9b,waldendorf-etal-2025-multilingual}, questions remain regarding their boundaries and limitations \cite{apple_cqf}. These unresolved concerns challenge the presumed universality of CQF as a filtering paradigm and require further investigation. Finally, while Wikipedia-style paraphrasing has been utilised to enhance low-quality data, prior works \cite{maini-etal-2024-rephrasing,su-etal-2025-nemotron} have employed it as a post-processing step applied solely to data that has already undergone filtering, and without explicit guidelines defining the extent of permissible modifications.

\section{CQF Pitfalls}
\subsection{Wikipedia-style rephrasing}\label{sec:wiki}

Our Wikipedia-style rephrasing experiment, illustrated in Figure \ref{fig:teaser}, comprises two distinct stages. In the first stage, we transform the original webpage content into a Wikipedia-style format, with a primary focus on restructuring the presentation to align with the conventional characteristics of Wikipedia articles. In the second stage, we apply CQF models to evaluate both the original and rephrased versions of the text, thereby generating filtering decisions for each variant. This two-stage methodology enables us to isolate the impact of formatting conventions on quality assessment.

In the study, we investigate three CQF models: the original \texttt{FineWeb-Edu}\footnote{\url{https:/hf.co/HuggingFaceFW/fineweb-edu-classifier}} \cite{fineweb} model and two models derived from Nemotron-CC, specifically \texttt{NemoCurator Mixtral}\footnote{\url{https://hf.co/nvidia/nemocurator-fineweb-mixtral-edu-classifier}} and \texttt{NemoCurator Nemotron}\footnote{\url{https://hf.co/nvidia/nemocurator-fineweb-nemotron-4-edu-classifier}} \cite{su-etal-2025-nemotron}. All three models utilise the Snowflake-Arctic-Embed-M embedding model as their foundation, a BERT-based model that contains 110 million parameters and supports a context window of 512 tokens \cite{snowflake}. Our analysis is based on a randomly sampled subset of 100,000 examples drawn from the FineWeb corpora \cite{fineweb}. To generate Wikipedia-style paragraphs, we employ the Qwen 2.5 72B Instruct\footnote{\url{https://hf.co/Qwen/Qwen2.5-72B-Instruct}} model \cite{qwen25}, which receives instructions to restructure the text webpage content while preserving the original information and modifying only the presentation format. The complete prompt used for this rephrasing task is available in Appendix \ref{sec:wiki_appendix} and the examples of raw and Wikipedia-style rephrased documents can be found in Appendix \ref{sec:wiki_examples_appendix}.

\begin{figure}[th!]
    \centering
    \includegraphics[width=\linewidth]{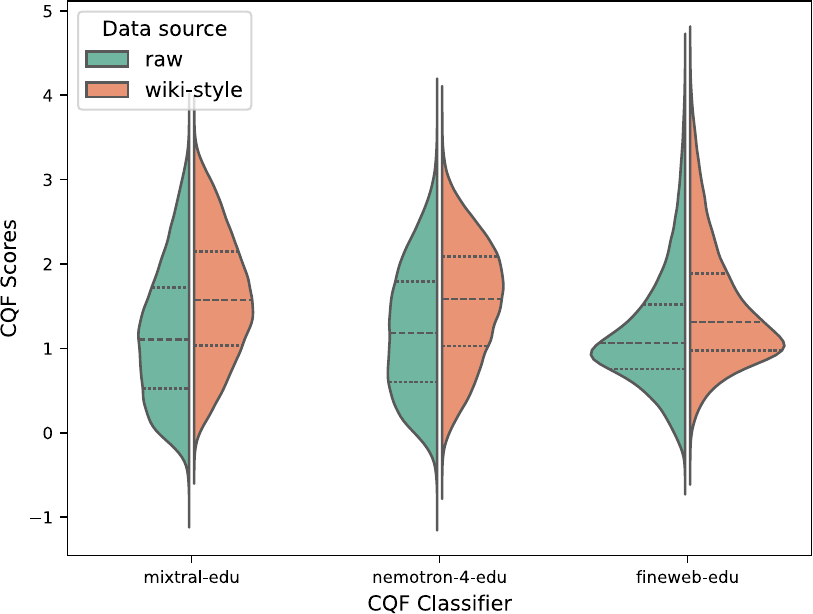}
    \caption{Wikipedia-style rephrasing impact on the CQF models' scores. The original documents (named ``raw'' - green, left part of the violin plots) score significantly lower for all evaluated models compared to their Wikipedia-style modified counterparts (orange, right).}
    \label{fig:wiki}
\end{figure}

Figure \ref{fig:wiki} presents the distributions of CQF scores assigned to both the original and rephrased versions of the text. All three models exhibit consistent behaviour, assigning higher scores to Wikipedia-style rephrasing. As shown in Table \ref{tab:wikiscores}, FineWeb-Edu demonstrates the highest robustness among the evaluated models, evidenced by the smallest disparity between scores for raw and Wikipedia-style content. Nevertheless, all models allow a substantial proportion of reformatted data to pass through their filters. Specifically, at a filtering threshold of 3, \texttt{NemoCurator Mixtral} admits over 7\% of the rephrased data that scores two or worse in the original form. The \texttt{NemoCurator Nemotron} and \texttt{FineWeb-Edu} admit 5\% and 6\%, respectively. Furthermore, even with a more rigorous threshold of 4, \texttt{FineWeb-Edu} fails to filter approximately 1\% of the Wikipedia-style data, which consistently receives elevated scores. Hence, while \texttt{FineWeb-Edu} performs best on average, it proves least effective in practice at eliminating false positives among the most highly-scored examples.

\begin{table}
    \centering
    \begin{tabular}{lcc}
    \toprule
    & \multicolumn{2}{c}{Mean score} \\
    CQF Model & raw & wiki-style \\
    \midrule
    \texttt{FineWeb-Edu} & 1.19 & 1.49 \\
    \texttt{NemoCurator Mixtral} & 1.17 & 1.60 \\
    \texttt{NemoCurator Nemotron} & 1.18 & 1.59 \\
    \bottomrule
    \end{tabular}
    \caption{CQF models' scores on the 100,000 examples without and with the Wikipedia-style rephrasing. While, on average, the FineWeb-Edu model is the most robust, as the difference between the original (raw) document and its rephrased version is the smallest, in practice, it is ``the worst'' choice (see Section \ref{sec:wiki} for details).}
    \label{tab:wikiscores}
\end{table}

\subsection{Domain sensitivity}\label{sec:domain}

\begin{figure*}[h!]
    \centering
    \includegraphics[width=\linewidth]{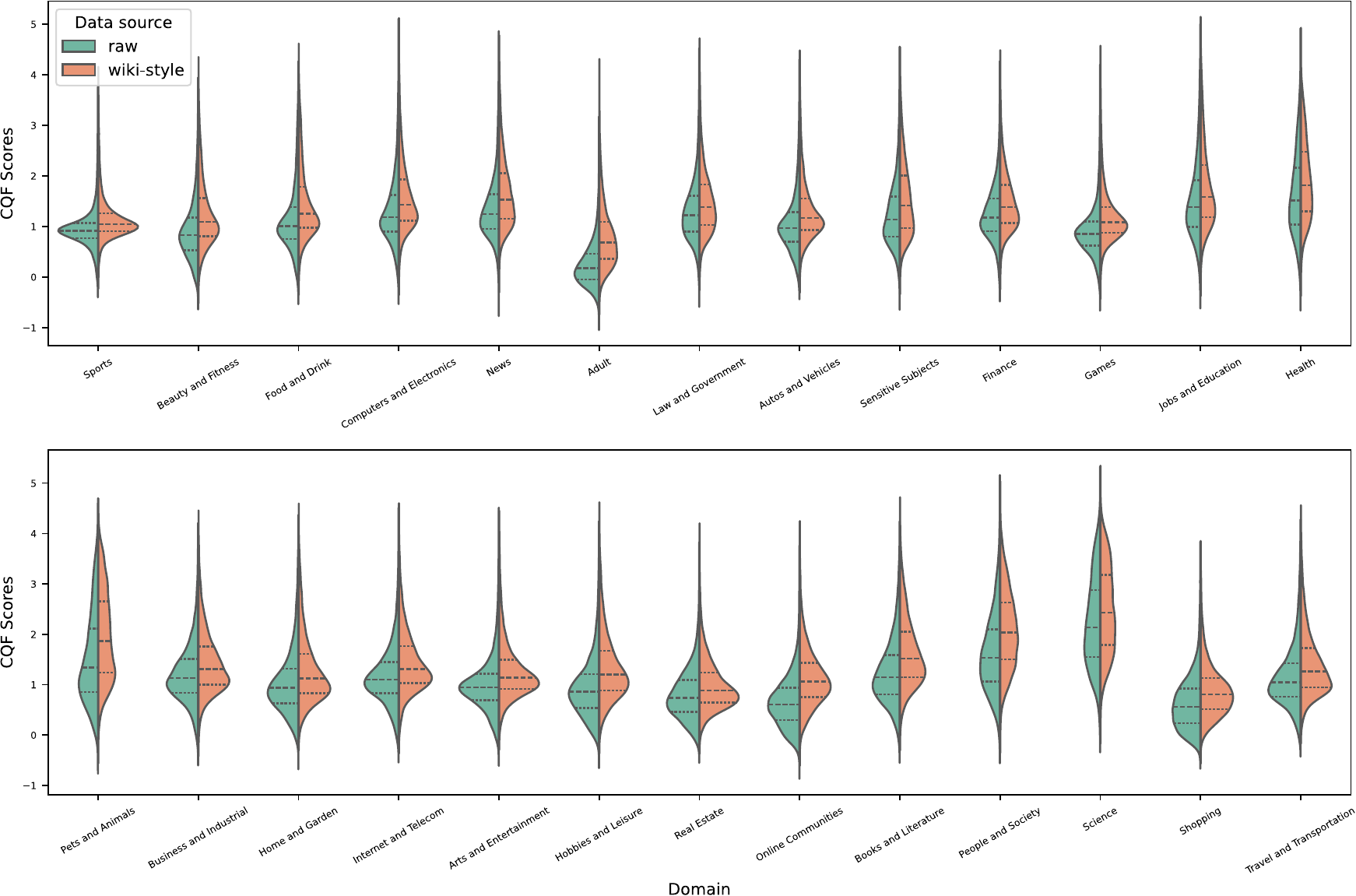}
    \caption{Wikipedia-style rephrasing impact on the FineWeb-Edu CQF model across domains. In each of 26 domains, the original documents score lower than their Wikipedia-style modified counterparts. Moreover, the distributions indicate that the "educational" approach implies a preference towards specific domains.}
    \label{fig:domain}
\end{figure*}

CQF models demonstrate bias not only toward Wikipedia-style articles but also toward particular content domains. For this phase of our analysis, we focus exclusively on the FineWeb-Edu model. We annotated the FineWeb corpus using Nvidia's domain classifier\footnote{\url{https://hf.co/nvidia/domain-classifier}} \cite{he2023debertav}, which assigns text to 26 distinct categories, and sampled 20,000 examples from each domain.

Figure \ref{fig:domain} presents the score distributions for both original and Wikipedia-style documents across these domains. The results reveal both anticipated behaviours and problematic patterns in CQF scoring. As expected, domains such as ``Science''—which intuitively should receive high-quality ratings—do indeed score among the highest. However, the rephrasing intervention consistently produces substantial score increases across all domains. In none of the 26 domains do the rephrased versions receive scores comparable to their original counterparts on average. The effect might reach substantial magnitude in certain domains; for instance, in ``Books and Literature'', the first quartile of Wikipedia-style rephrased documents nearly reaches the median score of the original documents.

Finally, comparing the distributions across domains, we can clearly see that "educational" filtering has a preference side effect. There is a definite over-representation of specific domains given a standard 3 or 4 threshold. While the result is not directly a pitfall of the CQF, it may end up being one if our downstream domain falls on the lower end of the model's preference. This finding aligns with the recent Nemotron-CLIMB \cite{climb} work, which highlights the limitations of CQF in specific domains and constructs pre-training corpora by searching for optimal pre-training data mixtures within predefined domains.

\section{Manual ``educational'' score annotation}
To determine whether the observed limitations of CQF models originate from the models themselves or from their training data, we revisited the LLM labelling process. Given the "educational" prompt of FineWeb-Edu (see Appendix \ref{sec:edu_appendix}), we instructed human annotators to evaluate webpages according to the same additive criteria employed in the original LLM-based annotation task. The annotation corpora include 100 documents, each of which was rated by three independent annotators\footnote{The group of human annotators consist of master's students from the NLP course.} according to the prompt. 
To curate the annotation corpora, we chose samples with the highest difference of the classifier score of the original and rephrased versions of the text. 
Then we 
sampled across all the possible scores to form a uniform distribution across the LLM-based scores.

The results reveal that a significant portion of the CQF models' shortcomings can be traced back to the LLM annotation process itself. 
Human annotators assigned "educational" scores that averaged 0.77 points lower than those generated by the LLM annotator (Llama 3.1 70B Instruct\footnote{\url{https://hf.co/meta-llama/Llama-3.1-70B-Instruct}} \cite{llama3}, consistent with the FineWeb approach). This discrepancy indicates that the CQF model, as a student model of LLMs, faces an even more challenging task in learning the meaning of rating educational content, given that its training data is inherently noisy, reflecting the biases and limitations of the teacher LLM.

\section{Conclusion}

In this paper, we identify a critical vulnerability inherent to Classifier-based Quality Filtering models. We demonstrate that simple Wikipedia-style reformatting of documents can systematically bypass filtering thresholds, enabling content that would otherwise be rejected to gain access into pre-training corpora. Furthermore, we establish that this phenomena manifest consistently across different domains, while confirming prior findings indicating that CQF models exhibit domain-dependent biases in their quality assessments. 
Our findings suggest that the ``educational'' quality scores assigned by the CQF teacher model may be inflated, providing a judgment that is overly optimistic regarding the actual value of the data.

Although CQF has undeniably contributed to the development of higher-performing LLMs, we hope that our findings will inform the evaluation and design of future CQF architectures. By presenting these limitations, we aim to mitigate the risk of introducing unintended biases and malicious data injections that could compromise downstream model behaviour and capabilities.

\section*{Acknowledgements}
This research was funded in whole by the National Science Centre, Poland 2023/49/N/ST6/02691. For the purpose of Open Access, the authors has applied a CC-BY public copyright licence to any Author Accepted Manuscript (AAM) version arising from this submission.
We gratefully acknowledge Polish high-performance computing infrastructure PLGrid (HPC Center: ACK Cyfronet AGH) for providing computer facilities and support within computational grant no. PLG/2025/018209.

\section*{Limitations}
Our analysis is based on automatic reformatting existing web corpora that must yield inaccuracies and include noisy outputs. Although scale and simplicity of the requested rephrasing should limit the variance of presented results, the exact numbers are imprecise and are an approximation of the actual problem degree.

\bibliography{custom,anthology-1,anthology-2}

\appendix

\section{Educational Prompt}
\label{sec:edu_appendix}
Figure \ref{fig:eduprompt} presents the FineWeb prompt used for ``educational'' LLM labelling. We provide the prompt for reference and the reader's convenience; the original prompt is available at \url{https://huggingface.co/HuggingFaceFW/fineweb-edu-classifier/blob/main/utils/prompt.txt}.

\begin{figure*}[t]
\begin{lstlisting}
Below is an extract from a web page. Evaluate whether the page has a high educational value and could be useful in an educational setting for teaching from primary school to grade school levels using the additive 5-point scoring system described below. Points are accumulated based on the satisfaction of each criterion:

- Add 1 point if the extract provides some basic information relevant to educational topics, even if it includes some irrelevant or non-academic content like advertisements and promotional material.
- Add another point if the extract addresses certain elements pertinent to education but does not align closely with educational standards. It might mix educational content with non-educational material, offering a superficial overview of potentially useful topics, or presenting information in a disorganized manner and incoherent writing style.
- Award a third point if the extract is appropriate for educational use and introduces key concepts relevant to school curricula. It is coherent though it may not be comprehensive or could include some extraneous information. It may resemble an introductory section of a textbook or a basic tutorial that is suitable for learning but has notable limitations like treating concepts that are too complex for grade school students.
- Grant a fourth point if the extract highly relevant and beneficial for educational purposes for a level not higher than grade school, exhibiting a clear and consistent writing style. It could be similar to a chapter from a textbook or a tutorial, offering substantial educational content, including exercises and solutions, with minimal irrelevant information, and the concepts aren't too advanced for grade school students. The content is coherent, focused, and valuable for structured learning.
- Bestow a fifth point if the extract is outstanding in its educational value, perfectly suited for teaching either at primary school or grade school. It follows detailed reasoning, the writing style is easy to follow and offers profound and thorough insights into the subject matter, devoid of any non-educational or complex content.

The extract:
<EXAMPLE>.

After examining the extract:
- Briefly justify your total score, up to 100 words.
- Conclude with the score using the format: "Educational score: <total points>"
\end{lstlisting}
\caption{Educational prompt listing.}
\label{fig:eduprompt}
\end{figure*}

\section{Wikipedia Rephrasing Prompt}
\label{sec:wiki_appendix}

Figure \ref{fig:rephrase} presents prompt used for the Wikipedia-style rephrasing.

\begin{figure*}[t]
\begin{lstlisting}
You are a Wikipedia-style rephraser. Your objective is to rephrase the following web page to imitate a Wikipedia article.

Web page:
```{web_page}```

Follow the rules below during rephrasing:
- Focus on containing all the facts from the document, even if they are not essential.
- Do not include new facts, concepts and overall new content.
- Keep the exact dates, locations, names and other entities.
- Outcome should differ only in terms of style and formatting.
- The output document should have a similar number of tokens (with a maximum 10% margin).
\end{lstlisting}
\caption{Wikipedia-style rephrasing prompt listing.}
\label{fig:rephrase}
\end{figure*}

\section{Wikipedia Rephrasing Examples}
\label{sec:wiki_examples_appendix}

Figures \ref{fig:rephrase_example1}, \ref{fig:rephrase_example2}, \ref{fig:rephrase_example3} present the Wikipedia-style rephrasing examples.

\begin{figure*}[t]

\small
Raw Document:
\begin{lstlisting}
Teach your students how to encrypt and decrypt binary messages as well as understand binary code. This resource includes a hidden January themed message which students need to decrypt using the key (ASCII Table). Students are also given the option to write their own encrypted binary message.
This is a perfect CS unplugged activity and can be used as an introduction to a lesson, for homework, for early finishers or even when you have no access to computers in the computer lab.
This is an ideal activity for the ISTE Standard of Computational Thinking/
Thanks to the following for all images:
Teachers ClipArt: www.edu-clips.com
Mike Rawls: http://www.teacherspayteachers.com/Store/Monster-Wrangler-Mike
\end{lstlisting}
Wikipedia-style rephrased document:
\begin{lstlisting}
### Binary Message Encryption and Decryption Activity

This educational resource is designed to teach students how to encrypt and decrypt binary messages and to understand binary code. The activity includes a hidden message with a January theme, which students must decrypt using the ASCII table. Additionally, students have the option to create their own encrypted binary messages.

This activity serves as an excellent "CS Unplugged" exercise and can be utilized in various educational settings, including as an introduction to a lesson, for homework, for early finishers, or when computers are unavailable in the computer lab. It aligns well with the ISTE Standard of Computational Thinking.

#### Credits
- **Images**: Provided by Teachers ClipArt (<https://www.edu-clips.com>) and Mike Rawls (<https://www.teacherspayteachers.com/Store/Monster-Wrangler-Mike>)

This resource is ideal for educators looking to introduce fundamental concepts of binary code and encryption in a practical and engaging manner.
\end{lstlisting}
\caption{Wikipedia-style rephrasing example \#1. The transformation reduced CQF score by 0.02 (4.17 vs 4.15).}
\label{fig:rephrase_example1}
\end{figure*}

\begin{figure*}[t]
\small
Raw Document:
\begin{lstlisting}
Ok, this is one of the most perfect cups of teas I've had. I made this yesterday, but ended up steeping it three times. Rather than have the chocolate filtered, I plucked the chips out of the filter and put them straight in my cup.
The recommendations suggest a heaping teaspoon per 12 oz. I had about 10 oz of water, so left off the "heaping."
I think because the chocolate chips were in the bottom on my cup, the sweetness was a little more concentrated towards the bottom of the cup. Most of the cup was subtly sweet and beautifully floral. Then, as I got to the bottom of the cup, the sweetness gets a bit stronger, but still beautiful. One of the best cups of comfort I've had in awhile.
One the second and third steep, I increased the steeping time to 3 and 4 minutes, respectively.
Flavors: Dark Chocolate, Rose, White Chocolate
\end{lstlisting}
Wikipedia-style rephrased document:
\begin{lstlisting}
== Preparation and Experience of a Notable Cup of Tea ==

A user recently prepared a cup of tea, which they described as one of the most perfect they have ever had. The tea was prepared and steeped three times over the course of a single day. Instead of using a filter for the chocolate, the chocolate chips were removed from the filter and placed directly into the cup.

The recommended measurement is a heaping teaspoon of tea per 12 ounces of water. However, since the user had approximately 10 ounces of water, they used a level teaspoon instead of a heaping one.

The placement of the chocolate chips at the bottom of the cup resulted in a concentration of sweetness towards the bottom. The majority of the cup was subtly sweet and wonderfully floral, with the sweetness becoming more pronounced towards the end of the cup. The user described it as one of the most comforting cups of tea they have had in a long time.

For the second and third steepings, the steeping times were increased to 3 and 4 minutes, respectively. The notable flavors experienced were dark chocolate, rose, and white chocolate.
\end{lstlisting}
\caption{Wikipedia-style rephrasing example \#2. The transformation increased CQF score by 2.24 (0.52 vs 2.76).}
\label{fig:rephrase_example2}
\end{figure*}

\begin{figure*}[t]
\small
Raw Document:
\begin{lstlisting}
I'm enjoying my Haiku series a bit too much lol. I would really love for you guys to try and reply my Haiku - use the comment box below.
You can make it a bit more fun by starting yours with the last line of mine :-) 
Looking forward to reading!
When the rain drops stop
Can you tell me what happens?
A rainbow or clouds...
\end{lstlisting}
Wikipedia-style rephrased document:
\begin{lstlisting}
### Haiku Series and Interactive Engagement

The author expresses their enjoyment of writing haiku and encourages readers to participate by composing their own haiku in the comments section. To make the activity more engaging, the author suggests that readers start their haiku with the last line of the author's haiku. The author looks forward to reading the responses.

#### Sample Haiku

When the rain drops stop,  
Can you tell me what happens?  
A rainbow or clouds...
\end{lstlisting}
\caption{Wikipedia-style rephrasing example \#3. The transformation increased CQF score by 1.83 (-0.04 vs 1.79).}
\label{fig:rephrase_example3}
\end{figure*}

\end{document}